\documentclass[runningheads]{llncs}
\usepackage{graphicx}
\usepackage{amsmath,amssymb} 
\usepackage{color}
\usepackage{mmstyles}
\usepackage{array}
\usepackage{multirow}

\newcommand{\tuple}[3]{$\langle$\emph{#1-#2-#3}$\rangle$}

\newcolumntype{P}[1]{>{\centering\arraybackslash}p{#1}}
\newcolumntype{M}[1]{>{\centering\arraybackslash}m{#1}}

\makeatletter
\def\hlinew#1{%
  \noalign{\ifnum0=`}\fi\hrule \@height #1 \futurelet
   \reserved@a\@xhline}
\makeatother

\usepackage[width=122mm,left=12mm,paperwidth=146mm,height=193mm,top=12mm,paperheight=217mm]{geometry}
\begin{document}
\pagestyle{headings}
\mainmatter

\title{Zoom-Net: Mining Deep Feature Interactions for \\ Visual Relationship Recognition} 

\titlerunning{ZoomNet}

\authorrunning{Guojun Yin \etal}

\author{Guojun Yin$^{1,3}$, Lu Sheng$^3$, Bin Liu$^1$, Nenghai Yu$^1$, Xiaogang Wang$^3$, \\ Jing Shao$^2$, Chen Change Loy$^4$ }

\institute{ $^1$University of Science and Technology of China,
Key Laboratory of Electromagnetic Space Information, the Chinese Academy of Sciences, $^2$SenseTime Group Limited, \\ $^3$CUHK-SenseTime Joint Lab, The Chinese University of Hong Kong,  \\ $^4$SenseTime-NTU Joint AI Research Centre, Nanyang Technological University \\
   \email{ gjyin@mail.ustc.edu.cn, \{flowice,ynh\}@ustc.edu.cn, ccloy@ieee.org, \{lsheng,xgwang\}@ee.cuhk.edu.hk, shaojing@sensetime.com}
}

\maketitle

\begin{abstract}
   Recognizing visual relationships $\langle$subject-predicate-object$\rangle$  among any pair of localized objects is pivotal for image understanding.
   Previous studies have shown remarkable progress in exploiting linguistic priors or external textual information to improve the performance.
   In this work, we investigate an orthogonal perspective based on feature interactions. We show that by encouraging deep message propagation and interactions between local object features and global predicate features, one can achieve compelling performance in recognizing complex relationships without using any linguistic priors. To this end, we present two new pooling cells to encourage feature interactions:~(i) Contrastive ROI Pooling Cell, which has a unique deROI pooling that inversely pools local object features to the corresponding area of global predicate features.~(ii) Pyramid ROI Pooling Cell, which broadcasts global predicate features to reinforce local object features.
   The two cells constitute a~\textit{Spatiality-Context-Appearance Module~(SCA-M)}, which can be further stacked consecutively to form our final~\textit{Zoom-Net}.
   We further shed light on how one could resolve ambiguous and noisy object and predicate annotations by Intra-Hierarchical trees~(IH-tree).
   Extensive experiments conducted on Visual Genome dataset~\cite{krishna2017vg} demonstrate the effectiveness of our feature-oriented approach compared to state-of-the-art methods~(Acc@$1$  $11.42\%$ from $8.16\%$~\cite{li2017vip}) that depend on explicit modeling of linguistic interactions. We further show that SCA-M can be incorporated seamlessly into existing approaches~\cite{li2017vip} to improve the performance by a large margin. The source code will be released on \url{https://github.com/gjyin91/ZoomNet}.
\end{abstract}

\section{Introduction}
\label{sec:intro}

\begin{figure*}[t]
\centering
\includegraphics[width=1\linewidth]{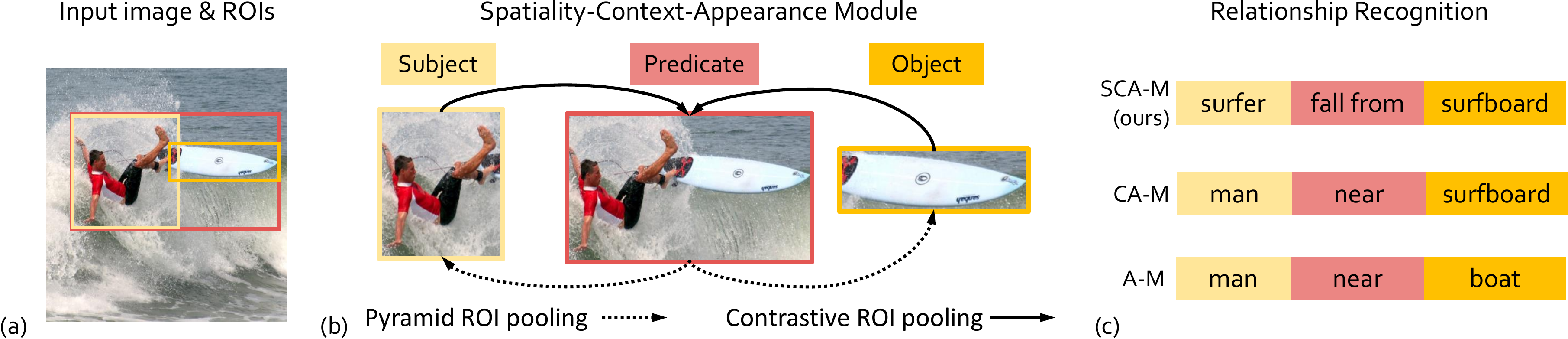}
\caption{Given an image~`\textit{surfer fall from surfboard}' and its region-of-interests~(ROI) in~(a), traditional methods without mining contextual interactions between object~(subject) and predicate~(\eg, Appearance Module~(A-M)) or ignoring spatial information~(\eg, Context-Appearance Module~(CA-M)) may fail in relationship recognition, as shown in the two bottom rows of~(c). The proposed Spatiality-Context-Appearance Module~(SCA-M) in~(b) permits global inter-object interaction and sharing of spatiality-aware contextual information, thus leading to a better recognition performance.
}
\label{fig:overview}
\end{figure*}

%
Visual relationship recognition~\cite{krishna2017vg,lu2016lp,sadeghi2011vp} aims at interpreting rich interactions between a pair of localized objects,~\ie,~performing tuple recognition in the form of \tuple{subject}{predicate}{object} as shown in Fig.~\ref{fig:overview}(a). 
The fundamental challenge of this task is to recognize various vaguely defined relationships given diverse spatial layouts of objects and complex inter-object interactions.
To complement visual-based recognition, a promising approach is to adopt a linguistic model and learn relationships between object and predicate labels from language. This strategy has been shown effective by many existing methods~\cite{li2017vip,lu2016lp,dai2017drnet,zhang2017vtranse,zhang2017ppr,zhuang2017cai}.
These language-based methods either apply statistical inference to the tuple label set,  establish a linguistic graph as the prior, or mine linguistic knowledge from external billion-scale textual data~(\eg,~Wikipedia).

In this paper, we explore a novel perspective beyond the linguistic-based paradigm. 
In particular, contemporary approaches typically recognize the tuple \tuple{subject}{predicate}{object} via separate convolutional neural network~(CNN) branches.
We believe that by enhancing message sharing and feature interactions among these branches, the participating objects and their visual relationship can be better recognized.
To this end, we formulate a new spatiality-aware contextual feature learning model, named as \textbf{Zoom-Net}. Differing from previous studies that learn appearance and spatial features separately\footnote{Their spatiality-streams simply apply the union region~\cite{sadeghi2011vp}, binary masks~\cite{dai2017drnet} or centroid coordinates~\cite{zhuang2017cai,zhang2017rlationpn,yu2017visual} as the abstraction of the spatial features.}, Zoom-Net propagates spatiality-aware object features to interact with the predicate features and broadcasts predicate features to reinforce the features of subject and object. 

The core of Zoom-Net is a \textbf{Spatiality-Context-Appearance Module}, abbreviated as \textbf{SCA-M}. It consists of two novel pooling cells that permit deep feature interactions between objects and predicates, as shown in Fig.~\ref{fig:overview}(b).
The first cell,~\textit{Contrastive ROI Pooling Cell}, facilitates predicate feature learning by inversely pooling object/subject features to a matching spatial context of predicate features via a unique deROI pooling. This allows all subject and object to fall on the same spatial `palette' for spatiality-aware feature learning.
The second cell is called~\textit{Pyramid ROI Pooling Cell}. It helps object/subject feature learning through broadcasting the predicate features to the corresponding object's/subject's spatial area.
Zoom-Net stacks multiple SCA-Ms consecutively in an end-to-end network that allows multi-scale bidirectional message passing among subject, predicate and object. 
As shown in Fig.~\ref{fig:overview}(c), the message sharing and feature interaction not only help recognize individual objects more accurately but also facilitate the learning of inter-object relation.

Another contribution of our work is an effective strategy of mitigating ambiguity and imbalanced data distribution in \tuple{subject}{predicate}{object} annotations. Specifically, we conduct our main experiments on the challenging Visual Genome~(VG) dataset~\cite{krishna2017vg}, which consists of over $5,319$ object categories, $1,957$ predicates, and $421,697$ relationship types.
The large-scale ambiguous categories and extremely imbalanced data distribution in VG dataset~(Tab.~\ref{tb:vrd_vs_vg},\ref{tb:vg_statistic}) prevent previous methods from predicting reliable relationships despite they succeed in the Visual Relationship Detection~(VRD) dataset~\cite{lu2016lp} with only $100$ object categories, $70$ predicates and $6,672$ relationships.
To alleviate the ambiguity and imbalanced data distribution in VG, we reformulate the conventional one-hot classification as a $n$-hot multi-class hierarchical recognition via a novel Intra-Hierarchical trees~(IH-trees) for each label set in the tuple \tuple{subject}{predicate}{object}.

\noindent\textbf{Contributions}. Our contributions are summarized as follows:

\noindent 1)~\textit{A general feature learning module that permits feature interactions} -
We introduce a novel SCA-M to mining intrinsic interactions between low-level spatial information and high-level semantical appearance features simultaneously. By stacking multiple SCA-Ms into a Zoom-Net, we achieve compelling results on VG dataset thanks to the multi-scale bidirectional message passing among subject, predicate and object.

\noindent 2)~\textit{Multi-class Intra-Hierarchical tree} -
To mitigate label ambiguity in large-scale datasets, we reformulate the visual relationship recognition problem to a multi-label recognition problem. The recognizability is enhanced by introducing an Intra-Hierarchical tree~(IH-tree) for the object and predicate categories, respectively. We show that IH-tree can benefit other existing methods as well.

\noindent 3)~\textit{Large-scale relationship recognition} -
Extensive experiments demonstrate the respective effectiveness of the proposed SCA-M and IH-tree, as well as their combination on the challenging large-scale VG dataset. 

It is noteworthy that the proposed method differs significantly from previous works as Zoom-Net neither models explicit nor implicit label-level interactions between \tuple{subject}{predicate}{object}. We show that feature-level interactions alone, which is enabled by SCA-M, can achieve state-of-the-art performance. We further demonstrate that previous state-of-the-arts~\cite{li2017vip} that are based on label-level interaction can benefit from the proposed SCA-M and IH-trees.

\section{Related work}

\noindent \textbf{Contextual Learning.} 
Contextual information has been employed in various tasks \cite{alexe2012searching,desai2011discriminative,gkioxari2015contextual,li2016human,park2010multiresolution,seco2004similarity,torralba2010using},~\eg,~object detection, segmentation, and retrieval.
For example, the visual features captured from a bank of object detectors are combined with  global features in~\cite{li2010object,choi2010exploiting}.
For both detection and segmentation, learning feature representations from a global view rather than the located object itself has been proven effective in~\cite{carreira2012object,li2011extracting,mottaghi2014role}.
Contextual feature learning for visual relationship recognition is little explored in previous works.

\noindent \textbf{Class Hierarchy.} 
In previous studies~\cite{hu2016learning,deng2012Hedging,ordonez2013large,deng2011hierarchical,deng2014large}, class hierarchy that encodes diverse label relations or structures is used to improve performances on classification and retrieval. 
For instance, Deng \etal~\cite{deng2012Hedging} improve large-scale visual recognition of object categories through forming a semantic hierarchy that consists of many levels of abstraction. 
While object categories can be clustered easily by their semantic similarity given the clean and explicit labels of objects, building a semantic hierarchy for visual relationship recognition can be more challenging due to noisy and ambiguous labels. Moreover, the semantic similarity between some phrases and prepositions such as~\textit{walking on a} versus~\textit{walks near the} is not directly measurable. In our paper, we employ the part-of-speech tagger toolkit to extract and normalize the keywords of these labels,~\eg~\textit{walk},~\textit{on} and~\textit{near}.


\noindent \textbf{Visual Relationship.} 
Recognizing visual relationship~\cite{sadeghi2011vp} has been shown beneficial to various tasks, including action recogntion~\cite{gkioxari2015contextual,delaitre2011learning}, pose estimation~\cite{desai2012detecting}, recognition and object detection~\cite{chen2013neil,rabinovich2007objects}, and scene graph generation~\cite{xu2017scene,li2017scene}.
Recent works~\cite{lu2016lp,dai2017drnet,zhang2017ppr,li2017scene,hu2017modular,liang2017recurrent,liangxiaodan2017vrl,peyre2017weakly,yatskar2016situation,zhuang2017care} show remarkable progress in visual relationship recognition, most of which focus on measuring linguistic relations with textual priors or language models.
The linguistic relations have been explored for object recognition~\cite{deng2014large,marszalek2007semantic,wang2009learning}, object detection~\cite{redmon2017yolo}, retrieval~\cite{schuster2015generating}, and caption generation~\cite{guadarrama2013youtube2text,karpathy2015deep,karpathy2014deep}.
%
%
%
Yu \etal~\cite{yu2017visual} employ billions of external textual data to distill useful knowledge for triplet \tuple{subject}{predicate}{object} learning.
These methods do not fully explore the potential of feature learning and feature-level message sharing for the problem of visual relationship recognition.
Li \etal~\cite{li2017vip} propose a message passing strategy to encourage feature sharing between features extracted from \tuple{subject}{predicate}{object}. However, the network does not capture the relative location of different objects thus it cannot capture valid contextual information between subject, predicate and object.


\begin{figure}[t]
\centering
\includegraphics[width=0.95\linewidth]{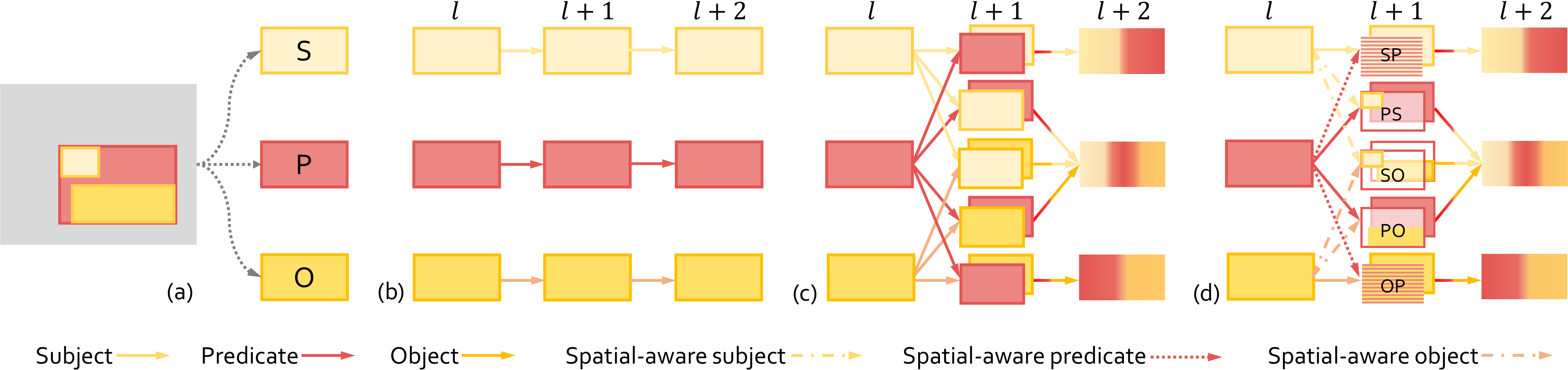}
\caption{(a) Given the ROI-pooled features of subject~(S), predicate~(P) and object~(O) from an input image,~(b) An Appearance module~(A-M) separately processes these features without any message passing,~(c) A Context-Appearance module~(CA-M) attempts to capture contextual information by directly fusing pairwise features. The proposed SCA-M in~(d) integrates the local and global contextual information in a spatiality-aware manner.
The SP/PS/SO/PO/OP features are combined by channel-wise concatenation. For instance, SP feature is the result of combining subject and predicate features.
}
\label{fig:module_compare}
\end{figure}

\section{Zoom-Net: Mining Deep Feature Interactions}
\label{sec:zoom_net}

We propose an end-to-end visual relationship recognition model that is capable of mining feature-level interactions. This is beyond just measuring the interactions among the triplet labels with additional linguistic priors, as what previous studies considered.

\subsection{Appearance, Context and Spatiality}
\label{subsec:app_vs_context_vs_spatiality}

As shown in Fig.~\ref{fig:module_compare}(a), given the ROI-pooled features of the subject, predicate and object, we consider a question: how to learn good features for both object~(subject) and predicate? We investigate three plausible modules as follows.

\noindent \textbf{Appearance Module.}
This module focuses on the intra-dependencies within each ROI,~\ie, the features of the subject, predicate and object branches are learned independently without any message passing.
We term this network structure as Appearance Module~(A-M), as shown in Fig.~\ref{fig:module_compare}(a).
No contextual and spatial information can be derived from such a module.

\noindent \textbf{Context-Appearance Module.}
The Context-Appearance Module~(CA-M)~\cite{li2017vip} directly fuses pairwise features among three branches, in which subject/object features absorb the contextual information from the predicate features, and predicate features also receive messages from both subject/object features, as shown in Fig.~\ref{fig:module_compare}(b).
Nonetheless, these features are concatenated regardless of their relative spatial layout in the original image. The incompatibility of scale and spatiality makes the fused features less optimal in capturing the required spatial and contextual information.

\noindent \textbf{Spatiality-Context-Appearance Module.}
The spatial configuration,~\eg,~the relative positions and sizes of subject and object, is not sufficiently represented in CA-M. 
To address this issue, we propose a Spatiality-Context-Appearance module~(\textbf{SCA-M}) as shown in Fig.~\ref{fig:module_compare}(c). It consists of two novel spatiality-aware feature alignment cells~(\ie,~\emph{Contrast ROI Pooling} and \emph{Pyramid ROI Pooling}) for message passing between different branches.
In comparison to CA-M, the proposed SCA-M reformulates the local and global information integration in a spatiality-aware manner, leading to superior capability in capturing spatial and contextual relationships between the features of \tuple{subject}{predicate}{object}.

\begin{figure*}[t]
\centering
\includegraphics[width=1\linewidth]{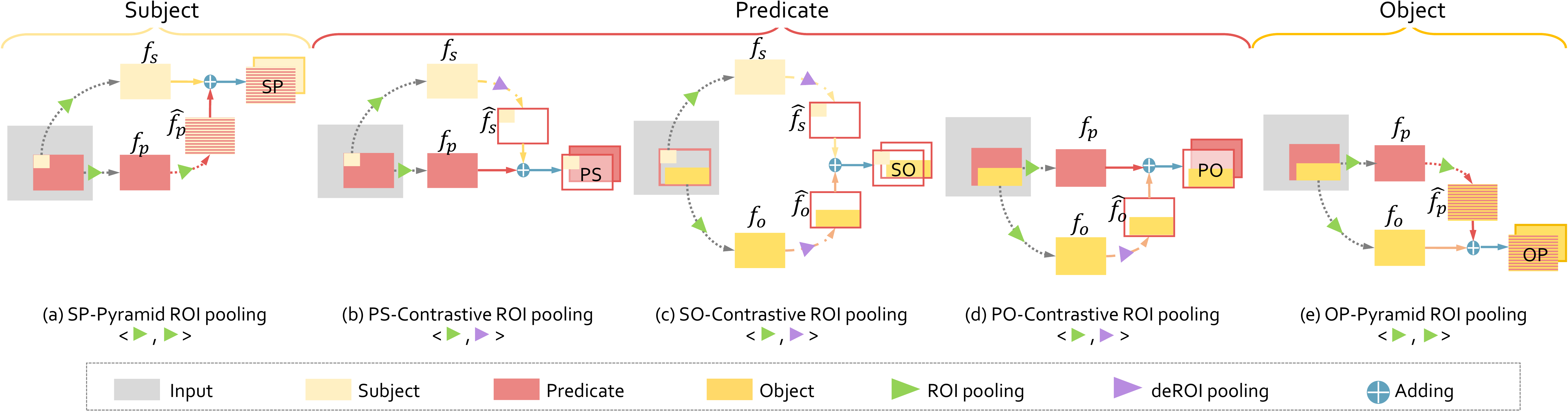}
\caption{The Spatiality-Context-Appearance Module~(SCA-M) hinges on two components:~(i) Contrastive ROI pooling~(b-d), denoted as $\langle$ROI, deROI$\rangle$, which propagates spatiality-aware features $\hat{f}_s, \hat{f}_o$ from~\textit{subject} and~\textit{object} into the spatial `palette' of~\textit{predicate} features $f_p$, and~(ii) Pyramid ROI pooling~(a,e), $\langle$ROI, ROI$\rangle$, which broadcasts the global~\textit{predicate} features $\hat{f}_p$ to local features $f_s, f_o$ in~\textit{subject} and~\textit{object} branches.
}
\label{fig:module}
\end{figure*}

\subsection{Spatiality-Context-Appearance Module~(SCA-M)}
\label{subsec:sca_module}

We denote the respective regions of interest~(ROIs) of the subject, predicate and object as $\mathcal{R}_s$, $\mathcal{R}_p$, and $\mathcal{R}_o$, where $\mathcal{R}_p$ is the union bounding box that tightly covers both the subject and object. 
The ROI-pooled features for these three ROIs are $\mathbf{f}_t, t\in\{s, p, o\}$, respectively.
In this section, we present the details of SCA-M. In particular, we discuss how Contrastive ROI Pooling and Pyramid ROI Pooling cells, the two elements in SCA-M, permit deep feature interactions between objects and predicates.

\noindent\textbf{Contrastive ROI Pooling}
denotes a pair of $\langle$ROI, deROI$\rangle$ operations that the object\footnote{Subject and object refer to the same concept, thus we only take object as the example for illustration.} features $\mathbf{f}_o$ are at first ROI pooled for extracting normalized local features, and then these features are deROI pooled back to the spatial palette of the predicate feature $\mathbf{f}_p$, so as to generate a spatiality-aware object feature $\hat{\mathbf{f}}_o$ with the same size as the predicate feature, as shown in Fig.~\ref{fig:module}(b) marked by the purple triangle. Note that the remaining region outside the relative object ROI in $\hat{\mathbf{f}}_o$ is set to $0$. 
The spatiality-resumed local feature $\hat{\mathbf{f}}_o$ can thus influence the respective regions in the global feature map $\mathbf{f}_p$.
In practice, the proposed deROI pooling can be considered as an inverse operation of the traditional ROI pooling~(green triangle in Fig.~\ref{fig:module}), which is analogous to the top-down deconvolution versus the bottom-up convolution.

There are three Contrastive ROI pooling cells presented in the SCA-M module to integrate the feature pairs \emph{subject-predicate}, \emph{subject-object} and \emph{predicate-object}, as shown in Fig.~\ref{fig:module}(b-d).
Followed by several convolutional layers, the features from subject and object are spatially fused into the predicate feature for enhanced representation capability.
The proposed $\langle$ROI, deROI$\rangle$ operations differ from conventional feature fusion operations~(channel-wise concatenation or summation). The latter would introduce scale incompatibility between local subject/object features and global predicate features, which could hamper feature learning in subsequent convolutional layers.

\noindent\textbf{Pyramid ROI Pooling}
denotes a pair of $\langle$ROI, ROI$\rangle$ operations that broadcasts the global predicate features to local features in the subject and object branches, as shown in Fig.~\ref{fig:module}(a) and~(e).
Specifically, with the help of ROI pooling unit, we first ROI-pool the features of predicate from the input region $\tilde{\mathcal{R}}$, which convey global contextual information of the region. Next, we perform a second ROI pooling on predicate features with the subject/object ROIs to further mine the contextual information from the global predicate feature region. The Pyramid ROI pooling thus provides multi-scale contexts to facilitate subject/object feature learning.

\begin{figure*}[t]
\centering
\includegraphics[width=1\linewidth]{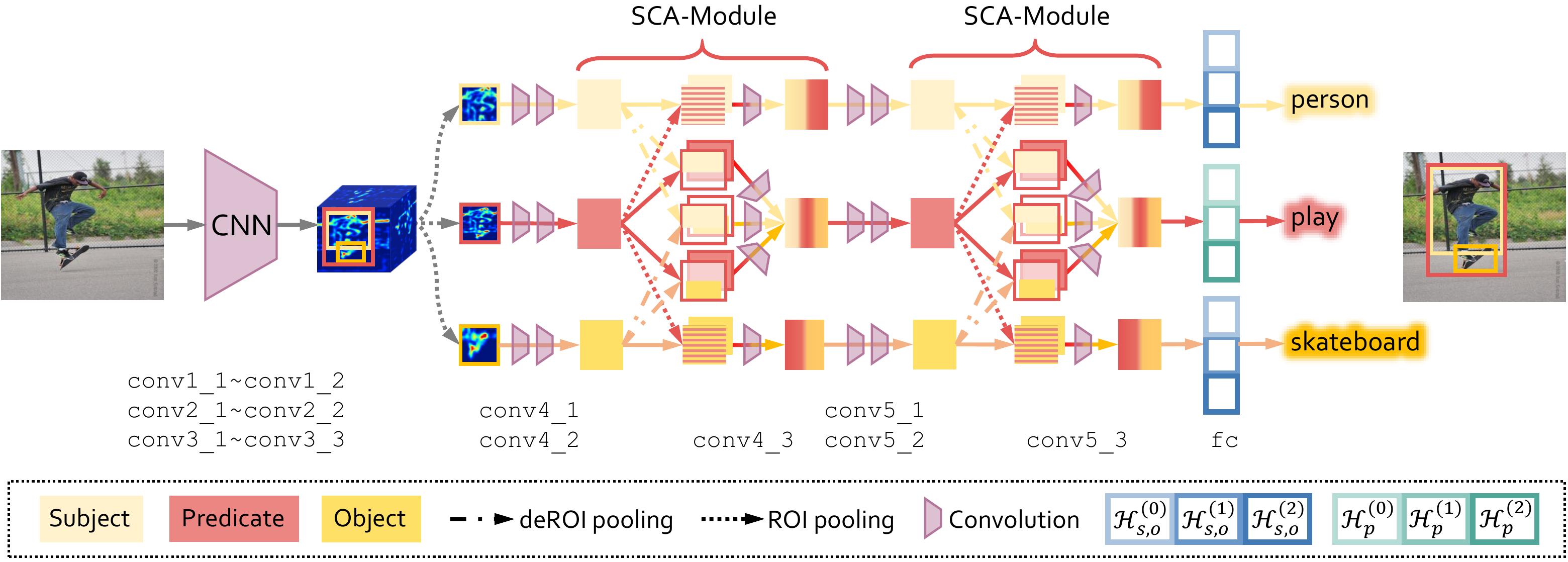}
\caption{The architecture of Zoom-Net. The subject~(in light yellow), predicate~(in red) and object~(in dark yellow) share the same feature extraction procedure in the lower layers, and are then ROI-pooled into three branches. Following each branch of pooled feature maps is two convolutional layers to learn appearance features which are then fed into two stacked SCA-Ms to further fuse multi-scale spatiality-aware contextual information across different branches. Three classifiers with intra-hierarchy structures are applied to the features obtained from each branch for visual relationship recognition.
}
\label{fig:framework}
\end{figure*}

\subsection{Zoom-Net: Stacked SCA-M}
\label{subsec:zoom_net}
By stacking multiple SCA-Ms, the proposed \textbf{Zoom-Net} is capable of capturing multi-scale feature interactions with dynamic contextual and spatial information aggregation.
It enables a reliable recognition of the visual relationship triplet \tuple{s}{p}{o}, where the predicate $p$ indicates the relationships~(\eg, spatiality, preposition, action and etc.) between a pair of localized subject $s$ and object $o$.

As visualized in Fig.~\ref{fig:framework}, we use a shared feature extractor with convolutional layers until \texttt{conv3\_3} to encode appearance features of different object categories.
By indicating the regions of interests~(ROIs) for subject, predicate and object, the associated features are ROI-pooled to the same spatial size and respectively fed into three branches.
The features in three branches are at first independently fed into two convolutional layers~(the \texttt{conv4\_1} and \texttt{conv4\_2} layers in VGG-16) for a further abstraction of their~\textit{appearance} features. 
Then these features are put into the first SCA-M to fuse~\textit{spatiality}-aware~\textit{contextual} information across different branches.
After receiving the interaction-augmented subject, predicate and object features from the first SCA-M, $\mathcal{M}^1_\text{SCA}$, we continue to convolve these features with another two appearance abstraction layers~(mimicking the structures of \texttt{conv5\_1} and \texttt{conv5\_2} layers in VGG-16) and then forward them to the second SCA-M, $\mathcal{M}^2_\text{SCA}$.
After this module, the multi-scale interaction-augmented features in each branch are fed into three fully connected layers \texttt{fc\_s}, \texttt{fc\_p} and \texttt{fc\_o} to classify subject, predicate and object, respectively.
%

\section{Hierarchical Relational Classification}
\label{sec:hierarchical_relation}

%
To thoroughly evaluate the proposed Zoom-Net, we adopt the Visual Genome~(VG) dataset\footnote{Extremely rare labels~(fewer than $10$ samples) were pruned for a valid evaluation.}~\cite{krishna2017vg} for its large scale and diverse relationships. 
Our goal is to understand the a much broader scope of relationships with a total number of $421,697$ relationship types~(as shown in Tab.~\ref{tb:vrd_vs_vg}), in comparison to the VRD dataset~\cite{lu2016lp} that focuses on only $6,672$ relationships. 
Recognizing relationships in VG is a non-trivial task due to several reasons:
%

\begin{table}[t]
\centering
\caption{Statistical comparison between VG dataset with VRD dataset. Objects, Predicates and Relationships are abbreviated to Obj., Pred. and Rel..}
\label{tb:vrd_vs_vg}
\footnotesize
\begin{tabular}{M{2.2cm}|M{1.5cm}|M{2.5cm}|M{1.2cm}|M{1.5cm}|M{2.0cm}} 
\hline
Datasets & \#Images&\#Rel. Instances&\#Obj.&\#Pred.&\#Rel. Types \\
\hline
VRD~\cite{lu2016lp} & 5,000 & 37,933 & 100 & 70 & 6,672 \\ 
\textbf{VG~\cite{krishna2017vg}} & \textbf{108,077} & \textbf{1,715,275} & \textbf{5,319} & \textbf{1,957} & \textbf{421,697} \\
\hline
\end{tabular}
\end{table}

\begin{table}[t]
\centering
\caption{The distribution of instance number per class of object and predicate, respectively. Both object and predicate have a long-tail distribution and only a few categories occur frequently.}
\label{tb:vg_statistic}
\footnotesize
\begin{tabular}{M{3.0cm}||M{1.8cm}M{1.8cm}M{2.0cm}M{2.1cm}} 
\hline
\#Ins.perClass&[10,100)&[100,500)&[500,1000)&[1000,$+\infty$)\\
\hline
\#Obj.Class & 3,919 & 892 & 209 & 299\\
\hline
\#Pred.Class & 1,501 & 313 & 55 & 88\\
\hline
\end{tabular}
\end{table}

\noindent (1)~\textit{Variety}~- 
There are a total of $5,319$ object categories and $1,957$ predicates, tens times than those available in the VRD dataset. 

\noindent (2)~\textit{Ambiguity}~- 
Some object categories share a similar appearance, and multiple predicates refer to the same relationship.

\noindent (3)~\textit{Imbalance}~- 
We observe long tail distributions both for objects and predicates.

To circumvent the aforementioned challenges, existing studies typically simplify the problem by manually removing a considerable portion of the data 
by frequency filtering or cleaning~\cite{li2017vip,dai2017drnet,zhang2017vtranse,yu2017visual}.
Nevertheless, infrequent labels like ``old man'' and ``white shirt'' contain common attributes like ``man'' and ``shirt'' and are unreasonable to be pruned.
Moreover, the flat label structure assumed by these methods is limited to describe the label space of the VG dataset with ambiguous and noisy labels.

To overcome the aforementioned issues, we propose a solution by establishing two Intra-Hierarchical trees~(IH-tree) for measuring intra-class correlation within object\footnote{Subject and object refer to the same term in this paper, thus we only take the object as the example for illustration.} and predicate, respectively. IH-tree builds a hierarchy of concepts that systematically groups rare, noisy and ambiguous labels together with those clearly defined labels.
Unlike existing works that regularize relationships across the triplet \tuple{$s$}{$p$}{$o$} by external linguistic priors, we only consider the intra-class correlation to independently regularize the occurrences of the object and predicate labels.
During end-to-end training, the network employs the weighted Intra-Hierarchical losses for visual relationship recognition as $\mathcal{L} = \alpha\mathcal{L}_s+\beta\mathcal{L}_p+\gamma\mathcal{L}_o$, where hyper-parameters $\alpha, \beta, \gamma$ balance the losses with respect to subject $\mathcal{L}_s$, predicate $\mathcal{L}_p$ and object $\mathcal{L}_o$. $\alpha=\beta=\gamma=1$ in our experiments.
We introduce IH-tree and the losses next.

\begin{figure*}[t]
\centering
\includegraphics[width=0.9\linewidth]{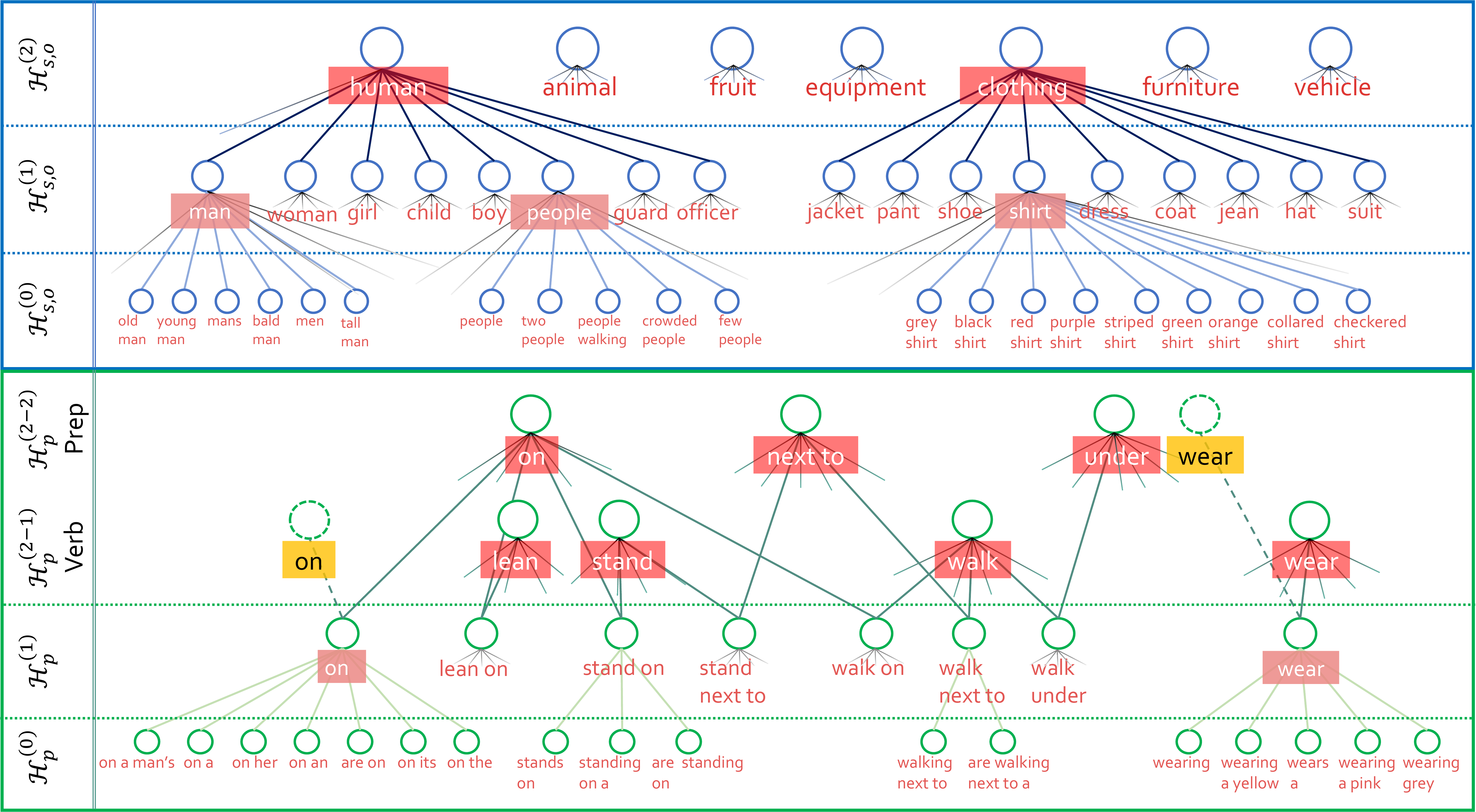}
\caption{An illustration of Intra-Hierarchical Tree. Both IH-trees for object~(top) and predicate~(bottom) start from the base layer $\mathcal{H}_{s,p,o}^{(0)}$ to a purified layer $\mathcal{H}_{s,p,o}^{(1)}$ but have a different construction in the third layer. The $\mathcal{H}_{o}^{(2)}$ clusters similar semantic concepts from $\mathcal{H}_{o}^{(1)}$, while the $\mathcal{H}_{p}^{(2)}$ separately cluster verb and preposition words from $\mathcal{H}_{p}^{(1)}$.
}
\label{fig:loss_hierarichy}
\end{figure*}

\subsection{Intra-Hierarchical Tree $\mathcal{H}_o$ for Object}

We build an IH-tree, $\mathcal{H}_o$, for object with a depth of three, where the base layer $\mathcal{H}_o^{(0)}$ consists of the raw object categories. 
%

\noindent(1) $\mathcal{H}_o^{(0)}\rightarrow\mathcal{H}_o^{(1)}$: $\mathcal{H}_o^{(1)}$ is extracted from $\mathcal{H}_o^{(0)}$ by pruning noisy labels with the same concept but different descriptive attributes or in different singular and plural forms.
We employ the part-of-speech tagger toolkit from NLTK
~\cite{bird2009nltk} and NLTK Lemmatizer to filter and normalize the noun keyword,~\eg, ``man'' from ``old man'', ``bald man'' and ``men''.
%

\noindent (2) $\mathcal{H}_o^{(1)}\rightarrow\mathcal{H}_o^{(2)}$: We observe that some labels have a close semantic correlation.
As shown in the top panel of Fig.~\ref{fig:loss_hierarichy}, labels with similar semantic concepts such as ``shirt'' and ``jacket'' are hyponyms of ``clothing'' and need to be distinguished from other semantic concepts like ``animal'' and ``vehicle''.
Therefore, we cluster labels in $\mathcal{H}_o^{(1)}$ to the third level $\mathcal{H}_o^{(2)}$ by semantical similarities computed by Leacock-Chodorow distance~\cite{seco2004similarity} from NLTK.
We find that a threshold of $0.65$ is well-suited for splitting semantic concepts.

The output of the subject/object branch is a concatenation of three independent \texttt{softmax} activated vectors corresponded to three hierarchical levels in the IH-tree.
The loss $\mathcal{L}_s$~($\mathcal{L}_o$) is thus a summation of three independent \texttt{softmax} losses with respect to these levels, encouraging the intra-level mutual label exclusion and inter-level label dependency.

\subsection{Intra-Hierarchical Tree $\mathcal{H}_p$ for Predicate}

The predicate IH-tree also has three hierarchy levels.
Different from the object IR-tree that only handles nouns, the predicate categories include various part-of-speech types,~\eg,~verb~(action) and preposition~(spatial position).
Even a single predicate label may contain multiple types,~\eg,~``are standing on'' and ``walking next to a''.

\noindent (1) $\mathcal{H}_p^{(0)}\rightarrow\mathcal{H}_p^{(1)}$:
Similar to $\mathcal{H}_o^{(1)}$, $\mathcal{H}_p^{(1)}$ is constructed aiming at extracting and normalizing keywords from predicates.
We retain the keywords and normalize tenses with respective to three main part-of-speech types,~\ie,~verb, preposition and adjective, and abandon other pointless and ambiguous words. As shown in the bottom panel of Fig.~\ref{fig:loss_hierarichy}, ``wears a'', ``wearing a yellow'' and ``wearing a pink'' are mapped to the same keyword ``wear''. 
%

\noindent (2) $\mathcal{H}_p^{(1)}\rightarrow\mathcal{H}_p^{(2)}$:
Different part-of-speech types own particular characteristics with various context representations, and hence a separate hierarchical structure for the verb~(action) and preposition~(spatial) is indispensable for better depiction. 
To this end, we construct $\mathcal{H}_p^{(2)}$ for verb and preposition label independently,~\ie,~$\mathcal{H}_p^{(2-1)}$ for action information and $\mathcal{H}_p^{(2-2)}$ for spatial configuration.
There are two cases in $\mathcal{H}_p^{(1)}$:~(a) the label is in the form of phrase that consists of both verb and preposition~(\eg~``stand on'' and ``walk next to'') and~(b) the label is a single word~(\eg,~``on'' and ``wear'').
For the first case, $\mathcal{H}_p^{(2-1)}$ extracts the verb words from the two phrases while $\mathcal{H}_p^{(2-2)}$ extracts the preposition words. It thus causes that a label might be simultaneously clustered into different partitions of $\mathcal{H}_p^{(2)}$.
If the label is a single word 
, it would be normally clustered into the corresponding part-of-speech but remained the same in the opposite part-of-speech, as shown with the dotted line in the bottom panel of Fig.~\ref{fig:loss_hierarichy}. 
The loss $\mathcal{L}_p$ is constructed similarly to that for the object.

\section{Experiments on Visual Genome~(VG) Dataset}
\label{sec:experiments_vg}

\noindent \textbf{Dataset.} 
We evaluate our method on the Visual Genome~(VG) dataset~(version 1.2). Each image is annotated with a triplet \tuple{subject}{predicate}{object}, where the subjects and objects are annotated with labels and bounding boxes while the predicates only have labels. The detailed statistics are stated in Sec.~\ref{sec:hierarchical_relation} and Tab.~\ref{tb:vrd_vs_vg},~\ref{tb:vg_statistic}.
We randomly split the VG dataset into training and testing set with a ratio of $8:2$. Note that both sets are guaranteed to have positive and negative samples from each object or predicate category.
The details of data preprocessing and the source code will be released.

\noindent \textbf{Evaluation Metrics.} 
%
\noindent (1)~\textit{Acc@$N$}. We adopt the~\textit{Accuracy} score as the major evaluation metric in our experiments. The metric is commonly used in traditional classification tasks. Specifically, we report the values of both Acc@$1$ and Acc@$5$ for~\textit{subject, predicate, object} and~\textit{relationship}, where the accuracy of~\textit{relationship} is calculated as the averaged accuracies of~\textit{subject, predicate} and~\textit{object}.

\noindent (2)~\textit{Rec@$N$}. Following~\cite{lu2016lp}, we use~\textit{Recall} as another metric so as to handle incomplete annotations. Rec@$N$ computes the ratio of the correct relationship instance that is covered in the top $N$ predictions per image. We report Rec@$50$ and Rec@$100$ in our experiments.
For a fair comparison, we follow~\cite{lu2016lp} to evaluate Rec@$N$ on three tasks,~\ie,~\textit{predicate recognition} where both the labels and bounding boxes of the subject and object are given;~\textit{phrase recognition} that takes a triplet as a union bounding box and predicts the triple labels;~\textit{relationship recognition}, which also outputs triple labels but evaluates separate bounding boxes of subject and object.
The recall performance is relative to the number of~\textit{predicate} per~\textit{subject-object} pair to be evaluated,~\ie, top $k$ predictions. In the experiments on VG dataset, we adopt top $k=100$ for evaluation.

\noindent \textbf{Training Details.} 
We use VGG$16$~\cite{simonyan2014vgg} pre-trained on ImageNet~\cite{deng2009imagenet} as the network backbone. The newly introduced layers are randomly initialized. We set the base learning rate as $0.001$ and fix the parameters from \texttt{conv1\_1} to \texttt{conv3\_3}. The implementations are based on Caffe~\cite{jia2014caffe}, and the networks are optimized via SGD.
The conventional feature fusion operations are implemented by channel-wise concatenation in SCA-M cells here.

\subsection{Ablation Study}
\label{subsec:ablation_study}

\noindent\textbf{SCA-Module.}
\label{subsubsec:exp_sca_module}
%
%
The advantage of Zoom-Net lies in its unique capability of learning spatiality-aware contextual information through the SCA-M.
To demonstrate the benefits of learning visual features with spatial-oriented and context-aided cues, we compare the recognition performance of Zoom-Net with a set of variants achieved by removing each individual cue step by step,~\ie.~the SCA-M without stacked structure, the CA-M that disregard the spatial layouts, and the vanilla A-M that does not perform message passing~(see Sec.~\ref{subsec:app_vs_context_vs_spatiality}). Their accuracy and recall scores are reported in Tab.~\ref{tb:exp_vg_comp_ablation}.

In comparison to vanilla A-M, both the CA-M and SCA-M obtain a significant improvement suggesting the importance of contextual information to individual subject, predicate, and object classification and their relationship recognition.
Note that contemporary CNNs have already shown a remarkable performance on subject and object classification,~\ie~it is not hard to recognize object via individual appearance information, and thus the gap~($4.96\%$) of subject is smaller than that of predicate~($12.25\%$) between A-M and SCA-M on Top-$1$ accuracy.
Not surprisingly, since the key inherent problem of relationship recognition is to learning the interactions between subject and object, the proposed SCA-M module exhibit a strong performance, thanks to its capability in capturing correlation between spatiality and semantic appearance cues among different object.
Its effectiveness can also be observed from qualitative comparisons in Fig.~\ref{fig:exp_qualitative_result}(a).

\begin{table*}[t]
\centering
\caption{Recognition performances~(Acc@$N$ and Rec@$N$) of Zoom-Net on VG dataset compared with~(i) three variants of SCA module, and~(ii) Zoom-Net discarding IH-trees. The best results are indicated in bold.}
\label{tb:exp_vg_comp_ablation}
\scriptsize
\begin{tabular}{ M{1cm}| M{1.5cm}| M{0.8cm}||M{1.2cm}| M{1.2cm} M{1cm} M{1cm} |M{1.7cm} M{1.6cm}}
\hlinew{1pt}
\multicolumn{2}{c|}{\textbf{Metrics}}& \textbf{@$N$} & Zoom-Net & SCA-M & CA-M & A-M & Zoom-Net w/o $\mathcal{H}^{(1,2)}$ & Zoom-Net w/o $\mathcal{H}^{(2)}$  \\
\hlinew{1pt}
\multirow{8}{*}{Acc.}  & \multirow{2}{*}{Subject}  &  1  & \textbf{38.94}  &  37.48 &  34.84 &  32.52 &  36.52 &  37.88 \\
 &   &  5  & \textbf{65.70} &  64.09 &  61.59 &  58.28 &  62.63 &  63.97 \\
 &  \multirow{2}{*}{Predicate}  &  1  & \textbf{48.73}  &  48.14 &  46.81 &  35.89 &  47.18 &  48.26 \\
 &   &  5  &  \textbf{77.64}  &  76.97 &  75.55 &  67.05 &  76.43 &  77.18 \\
 &  \multirow{2}{*}{Object}  &  1  & \textbf{45.09}  &  44.13 &  42.66 &  41.39 &  42.52 &  43.67 \\
 &   &  5  & \textbf{71.69}  &  70.64 &  69.55 &  67.99 &  69.33 &  70.35 \\
 &  \multirow{2}{*}{Relationship}  &  1  &  \textbf{11.42}  &  10.51 &  9.46 &  6.39 &  9.92 &  10.76 \\
 &   &  5  &  \textbf{22.80} &  21.31 &  19.70 &  14.06 &  20.44 &  22.08 \\
\hline
\multirow{6}{*}{Rec.}  &  \multirow{2}{*}{Predicate}  &  50  &  \textbf{67.25}  &  66.54 &  65.07 &  53.94 &  65.84 &  66.73 \\
 &   &  100  &  \textbf{77.51}  &  76.92 &  75.45 &  66.53 &  76.30 &  77.16 \\
 & \multirow{2}{*}{Relationship}  &  50  & \textbf{19.97}  &  18.60 &  17.14 &  12.23 &  17.78 &  18.92 \\
 &   &  100  &  \textbf{25.07}  &  23.51 &  21.63 &  15.86 &  22.53 &  23.88 \\
 &  \multirow{2}{*}{Phrase}  &  50  &  \textbf{20.84}  &  19.55 &  18.12 &  13.05 &  18.65 &  19.78 \\
 &   &  100  &  \textbf{26.16}  &  24.70 &  22.85 &  16.92 &  23.62 &  24.96 \\
\hlinew{1pt}
\end{tabular}
\end{table*}

\noindent\textbf{Intra-Hierarchical Tree.}
\label{subsubsec:exp_hr_tree}
%
We use the two auxiliary levels of hierarchical labels $\mathcal{H}^{(1)}$ and $\mathcal{H}^{(2)}$ to facilitate the prediction of the raw ground truth labels $\mathcal{H}^{(0)}$ for the subject, predicate and object, respectively.
Here we show that by involving hierarchical structures to semantically cluster ambiguous and noisy labels, the recognition performance \wrt the raw labels of the subject, predicate, object as well as their relationships are all boosted, as shown in Tab.~\ref{tb:exp_vg_comp_ablation}.
Discarding one of two levels in IH-tree clearly hamper the performance,~\ie,~Zoom-Net without IH-tree experiences a drop of around $1\%\sim 4\%$ on different metrics. It reveals that intra-hierarchy structures do provide beneficial information to improve the recognition robustness. 
Besides, Fig.~\ref{fig:exp_qualitative_result}(b) shows the Top-$5$ triple relationship prediction results of Zoom-Net with and without IH-trees.
The novel design of the hierarchical label structure help resolves data ambiguity for both on object and predicate.
For example, thanks to the hierarchy level $\mathcal{H}^{(1)}$ introduced in Sec.~\ref{sec:hierarchical_relation}, the predicates related to ``wear''~(\eg,~``wearing'' and ``wears'') can be ranked in top predictions.
Another example shows the contribution of $\mathcal{H}^{(2)}$ designed for semantic label clustering,~\eg~``sitting in'', which is grouped in the same cluster of the ground truth ``in'', also appears in top ranking results.

\subsection{Comparison with State-of-the-Art Methods}
\label{subsec:exp_comp_vg}

\begin{figure*}[t]
\centering
\includegraphics[width=1\linewidth]{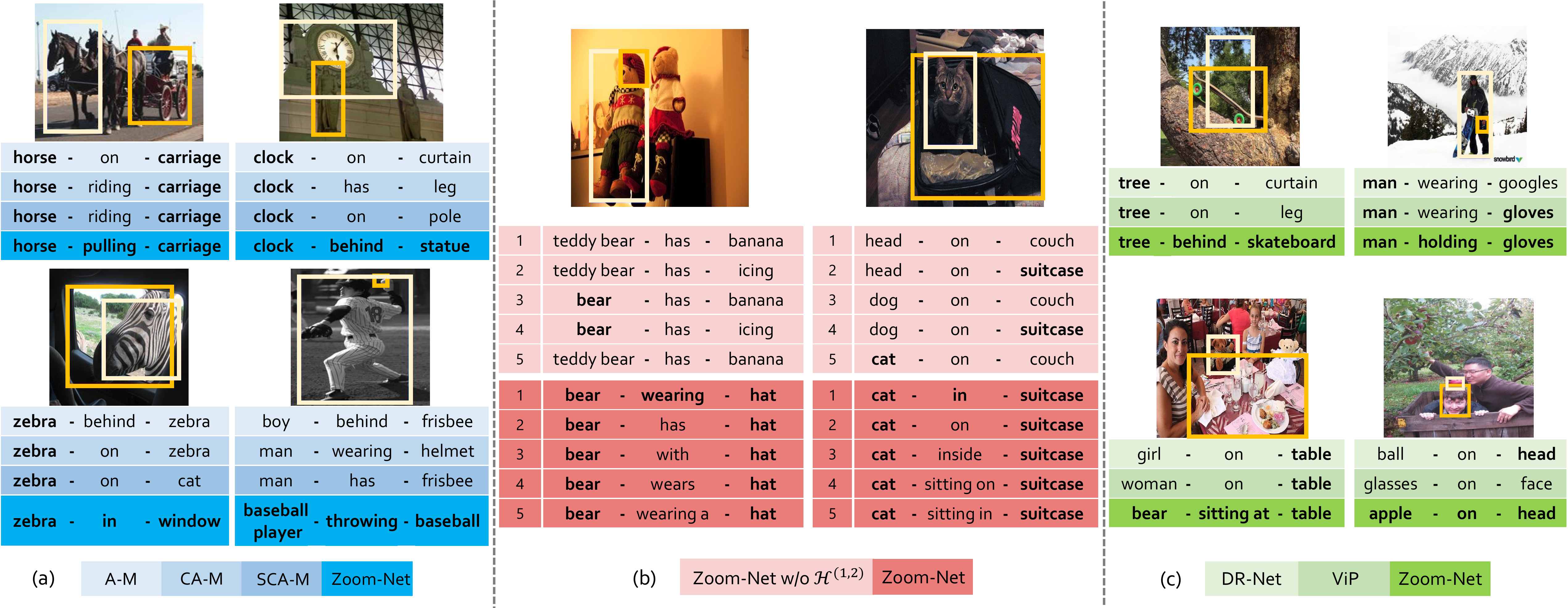}
\caption{Qualitative results on VG dataset.~(a) Comparison results with the variants of different module configurations.~(b) Results by discarding IH-trees.~(c) Comparison between Zoom-Net with state-of-the-art methods.~(a) and~(c) show Top-$1$ prediction results while~(b) provides Top-5 results for each method. The ground truth are in bold.
}
\label{fig:exp_qualitative_result}
\end{figure*}

We summarize the comparative results on VG in Tab.~\ref{tb:exp_vg_comp_methods} with two recent state of the arts~\cite{dai2017drnet,li2017vip}.
For a fair comparison, we implement both methods with the VGG-16 as the network backbone.
The proposed Zoom-Net significantly outperforms these methods, quantitatively and qualitatively.
Qualitative results are shown in the first row of Fig.~\ref{fig:exp_qualitative_result}(c).
DR-Net~\cite{dai2017drnet} exploits binary dual masks as the spatial configuration in feature learning and therefore loses the critical interaction between visual context and spatial information.
ViP~\cite{li2017vip} focuses on learning label interaction by proposing a phrase-guided message passing structure. Additionally, the method tries to capture contextual information by passing messages across triple branches before ROI pooling and thus fail to explore in-depth spatiality-aware feature representations.

\begin{table*}[t]
\centering
\caption{Recognition performances~(Acc@$N$ and Rec@$N$) of Zoom-Net on VG dataset compared with the state-of-the-art methods. Results in bold font are the best by a single model, while the underlined results indicate the best performance of a combined model that incorporates the proposed modules into other state-of-the-art architectures. }
\label{tb:exp_vg_comp_methods}
\scriptsize
\begin{tabular}{ M{0.7cm}| M{1.5cm}| M{0.8cm}||M{1.2cm}| M{1.3cm} M{1.1cm}|M{1.3cm} M{1.4cm} M{1.8cm}}
\hlinew{1.1pt}
\multicolumn{2}{c|}{\textbf{Metrics}}& \textbf{@$N$}  &  Zoom-Net  &  DR-Net\cite{dai2017drnet}  &  ViP\cite{li2017vip}  &  ViP+SCA-M  &  ViP+IH-tree &   ViP+SCA-M +IH-tree \\
\hlinew{1.1pt}
\multirow{8}{*}{Acc.}  & \multirow{2}{*}{Subject}  &  1  &  \textbf{38.94}  &  30.10 & 31.10 &~\textit{37.13}  &~\textit{34.36}  &~\textit{\underline{38.78}}  \\
 &   &  5  &  \textbf{65.70} & 55.46 & 57.33 &~\textit{ 63.61}  & ~\textit{61.03}  & ~\textit{\underline{65.69}}  \\
 &  \multirow{2}{*}{Predicate}  &  1  &  \textbf{48.73} & 44.14 & 45.17 &~\textit{48.40}  &~\textit{46.54}  &~\textit{\underline{49.07}}  \\
 &   &  5  &  \textbf{77.64} & 71.67 & 74.26 &~\textit{77.28}  &~\textit{75.30}  &~\textit{\underline{78.07}}  \\
 &  \multirow{2}{*}{Object}  &  1  &  \textbf{45.09} & 37.91 & 39.18 &~\textit{43.09}  &~\textit{43.18}  &~\textit{\underline{44.96}}  \\
 &   &  5  &  \textbf{71.69} & 64.30 & 65.68 &~\textit{69.93}  &~\textit{69.48}  &~\textit{\underline{71.58}}  \\
 &  \multirow{2}{*}{Relationship}  &  1  &  \textbf{ 11.42}  & 6.69 & 8.16 &~\textit{10.65}  &~\textit{9.97}  &~\textit{\underline{11.79} } \\
 &   &  5  &  \textbf{22.80}  & 13.11 & 17.01 &~\textit{21.63}  &~\textit{20.40}  &~\textit{\underline{23.28}}  \\
\hline
\multirow{6}{*}{Rec.}  &  \multirow{2}{*}{Predicate}  &  50  &  \textbf{ 67.25}  & 62.05 & 63.44 &~\textit{66.87}  &~\textit{64.80} &~\textit{\underline{67.63}}  \\
 &   &  100  &  \textbf{ 77.51}  & 71.96 & 74.15 &~\textit{77.22} &~\textit{75.29} &~\textit{\underline{77.89 }}\\
 & \multirow{2}{*}{Relationship}  &  50  &  \textbf{19.97}  & 12.56 & 14.78 &~\textit{18.73} &~\textit{17.76} &~\textit{\underline{20.41 }}\\
 &   &  100  &  \textbf{ 25.07}  & 16.06 & 18.85 &~\textit{23.67} &~\textit{22.35 }&~\textit{\underline{25.55}} \\
 &  \multirow{2}{*}{Phrase}  &  50  &  \textbf{ 20.84}  & 13.51 & 15.70 &~\textit{19.61} &~\textit{18.72} &~\textit{\underline{21.31}} \\
 &   &  100  &  \textbf{26.16}  & 17.23 & 19.96 &~\textit{24.70} &~\textit{23.50} &~\textit{\underline{26.66}} \\
\hlinew{1.1pt}
\end{tabular}
\end{table*}

\noindent\textbf{Transferable SCA-M Module and IH-Tree.}
We further demonstrate the effectiveness of the proposed SCA-M module in capturing spatiality, context and appearance visual cues, and IH-trees for resolving ambiguous annotations, by plugging them into architectures of existing works.
Here, we take the network of ViP~\cite{li2017vip} as the backbone for its end-to-end training scheme and state-of-the-art results~(Tab.~\ref{tb:exp_vg_comp_methods}).
We compare three configurations,~\ie,~\textit{ViP+SCA-M},~\textit{ViP+IH-tree} and~\textit{ViP+SCA-M+IH-tree}.
For a fair comparison, the ViP is modified by replacing the targeted components with SCA-M or IH-tree but with other components fixed.
As shown in Tab.~\ref{tb:exp_vg_comp_methods}, the performance of ViP is improved by a considerable margin on all evaluation metrics after applying our SCA-M~(\ie~\textit{ViP+SCA-M}). The results again suggest the superiority of the proposed spatiality-aware feature representations to that of ViP.
Note that the overall performance by adding both stacked SCA module and IH-tree~(\ie,~\textit{ViP+SCA-M+IH-tree}) surpasses that of ViP itself. The ViP designs a phrase-guided message passing structure to learn textual connections among \tuple{subject}{predicate}{object} at label-level. On the contrary, we concentrate more on capturing contextual connections among \tuple{subject}{predicate}{object} at feature-level. Therefore, it's not surprising that a combination of these two aspects can provide a better result.

\section{Comparisons on Visual Relationship Dataset~(VRD)}
\label{sec:exp_comp_vrd}
{
\noindent\textbf{Settings.}
We further quantitatively compare the performance of the proposed method with previous state of the arts on the Visual Relationship Dataset (VRD)~\cite{lu2016lp}. The following comparisons keep the same settings as the prior arts.
VRD dataset is widely used for its clean and accurate annotations, although it is much smaller and simpler than VG dataset as shown in Tab.~\ref{tb:vrd_vs_vg}.
Since VRD has a clean annotation, we fine-tune the construction of IH-tree by removing the $\mathcal{H}_o^{(1)}$ and $\mathcal{H}_p^{(1)}$, which aim at reducing data ambiguity and noise in VG~(details in Sec.~\ref{sec:hierarchical_relation}).
For a fair comparison, object proposals are generated by RPN~\cite{fastrcnn} here and we use triplet NMS to remove redundant triplet candidates following the setting in~\cite{li2017vip} due to its excellent performance.

\noindent\textbf{Evaluation metrics.}
We follow~\cite{dai2017drnet,yu2017visual} to report Recall@$50$ and Recall@$100$ when $k=70$. The IoU between the predicted bounding boxes and the ground truth is required above $0.5$ here. In addition, some previous works used $k=1$ for evaluation and thus we report our results with $k=1$ as well to compare these previous methods under the same conditions.

\begin{table}[t]
\centering
\caption{Comparisons with the referenced methods on VRD dataset. Results in bold indicate the best performance while the underlined results represent the next best. * marks the results of LK without knowledge distillation. ** marks the results of LK with knowledge distillation including large-scale external Wikipedia data.}
\label{tb:exp_vrd_comp}
\scriptsize
\begin{tabular}{M{1.0cm}| M{2.5cm}|M{1.3cm} M{1.3cm} M{1.3cm} M{1.3cm} M{1.3cm} M{1.3cm}} 
\hlinew{1.1pt}
\multirow{2}{*}{$k$} & \multirow{2}{*}{Methods} & \multicolumn{2}{c}{Predicate} & \multicolumn{2}{c}{Relationship} & \multicolumn{2}{c}{Phrase} \\
& & Rec@50 & Rec@100 & Rec@50 & Rec@100 & Rec@50 & Rec@100 \\
\hline
\multirow{11}{*}{$k=1$} & LP\cite{lu2016lp} & 47.87 & 47.87 & 13.86 & 14.70 & 16.17 & 17.03 \\
& VTransE\cite{zhang2017vtranse}& 44.76& 44.76& 14.07& 15.20& 19.42 & 22.42\\
& VRL\cite{liangxiaodan2017vrl}& -& -& {18.19} & {20.79} & 21.37& 22.60 \\
& PPRFCN\cite{zhang2017ppr}& 47.43& 47.43& 14.41& 15.72& 19.62& 23.15 \\
& SA-Full\cite{peyre2017weakly} & 50.40& 50.40 & 14.90 & 16.10 & 16.70 & 18.10 \\
& LK\cite{yu2017visual}* & 47.50 & 47.50 & 16.57& 17.69& 19.15& 19.98\\
& LK\cite{yu2017visual}** & \underline{55.16} & \underline{55.16} & \underline{19.17} & 21.34 & 23.14 & 24.03 \\
& ViP~\cite{li2017vip} & -& -& 17.32& 20.01& {22.78} & 27.91 \\
& CAI\cite{zhuang2017cai}& 53.59 & 53.59 & 15.63 & 17.39 & 17.60 & 19.24 \\
\cline{2-8}
& Zoom-Net & 50.69 & 50.69 & 18.92 & \underline{21.41} & \underline{24.82} & \underline{28.09} \\
&~\textit{CAI + SCA-M} &~\textit{\textbf{55.98}}&~\textit{\textbf{55.98}} &~\textit{\textbf{19.54}} &~\textit{\textbf{22.39}} &~\textit{\textbf{25.21}}&\textit{\textbf{28.89}}\\
\hline
\hline
\multirow{5}{*}{$k=70$} & LK[\cite{yu2017visual}* & 74.98& {86.97}& {20.12}& \underline{28.94} & {22.59}& {25.54} \\
& LK\cite{yu2017visual}** & \underline{85.64} & \textbf{94.65} & \textbf{22.68} & \textbf{31.89} & 26.32 & 29.43 \\
& DR-Net~\cite{dai2017drnet} & 80.78 & 81.90& 17.73& 20.88  & 19.93& 23.45 \\\cline{2-8}
& Zoom-Net & 84.25 & 90.59 & 21.37 & {27.30} & \underline{29.05} & \underline{37.34}\\
&~\textit{CAI + SCA-M} & ~\textit{\textbf{89.03}}&~\textit{\underline{94.56}} &~\textit{\underline{22.34}} &~\textit{28.52} &~\textit{\textbf{29.64}} &~\textit{\textbf{38.39}}\\
\hlinew{1.1pt}
\end{tabular}
\end{table}

\noindent\textbf{Results.}
The results listed in Tab.~\ref{tb:exp_vrd_comp} show that the proposed Zoom-Net outperforms the state-of-the-art methods by significant gains on almost all the evaluation metrics
\footnote{Note that Yu~\etal~\cite{yu2017visual} take external Wikipedia data with around $4$ billion and $450$ million sentences to distill linguistic knowledge 
for modeling the tuple correlation from label-aspect. 
It's not surprising to achieve a superior performance. In this experiment, we only compare with the results~\cite{yu2017visual} without knowledge distillation.}. 
In comparison to previous state-of-the-art approaches, Zoom-Net improves the recall of predicate prediction by $3.47\%$ Rec@$50$ and $3.62\%$ Rec@$100$ when $k=70$. Besides, the Rec@50 on relationship and phrase prediction tasks are increased by $1.25\%$ and $6.46\%$, respectively. 
Note that the result of~\textit{predicate}~($k=1$) only achieves comparable performance with some prior arts~\cite{zhuang2017cai,peyre2017weakly,liangxiaodan2017vrl,yu2017visual} since these methods use the groundtruth of~\textit{subject} and~\textit{object} and only predict~\textit{predicate} while our method predicts~\textit{subject, predicate, object} together. 

Among all prior arts designed without external data, CAI~\cite{zhuang2017cai} has achieved the best performances on predicate prediction~($53.59\%$ Rec@$50$) by designing a context-aware interaction recognition framework to encode the labels into semantic space. To demonstrate the effectiveness and robustness of the proposed SCA-M in feature representation, we replace the visual feature representation in CAI~\cite{zhuang2017cai} with our SCA-M
~(\ie~\textit{CAI + SCA-M}). The performance improvements are significant as shown in Tab.~\ref{tb:exp_vrd_comp} due to the better visual feature learned,~\eg, predicate Rec@50 is increased by $2.39\%$ compared  to~\cite{zhuang2017cai}. 
In addition, with neither language priors, linguistic models nor external textual data, the proposed method can still achieve the state-of-the-art performance on most of the evaluation metrics, thanks to its superior feature representations. 
}

\section{Conclusion}
\label{sec:conclusion}

We have presented an innovative framework Zoom-Net for visual relationship recognition, concentrating on feature learning with a novel Spatiality-Context-Appearance module~(SCA-M). The unique design of SCA-M, which contains the proposed Contrastive ROI Pooling and Pyramid ROI Pooling Cells benefits the learning of spatiality-aware contextual feature representation. We further designed the Intra-Hierarchical tree~(IH-tree) to model intra-class correlations for handling ambiguous and noisy labels. Zoom-Net achieves the state-of-the-art performance on both VG and VRD datasets. We demonstrated the superiority and transferability of each component of Zoom-Net. It is interesting to explore the notion of feature interactions in other applications such as image retrieval and image caption generation.

\subsubsection{Acknowledgment}
This work is supported in part by the National Natural Science Foundation of China (Grant No. 61371192), the Key Laboratory Foundation of the Chinese Academy of Sciences (CXJJ-17S044) and the Fundamental Research Funds for the Central Universities (WK2100330002, WK3480000005), in part
by SenseTime Group Limited, the General Research Fund sponsored by the Research Grants Council of Hong Kong (Nos. CUHK14213616, CUHK14206114, CUHK14205615, CUHK14203015, CUHK14239816, CUHK419412, CUHK14207-814, CUHK14208417, CUHK14202217), the Hong Kong Innovation and Technology Support Program (No.ITS/121/15FX).

\bibliographystyle{splncs}
\bibliography{egbib}

\clearpage

\section{Appendix}
\label{sec:appendix}

\subsection{Intra-Hierarchical Trees}
\label{sec:ihtree}
\begin{figure}
\centering
\includegraphics[width=0.49\columnwidth]{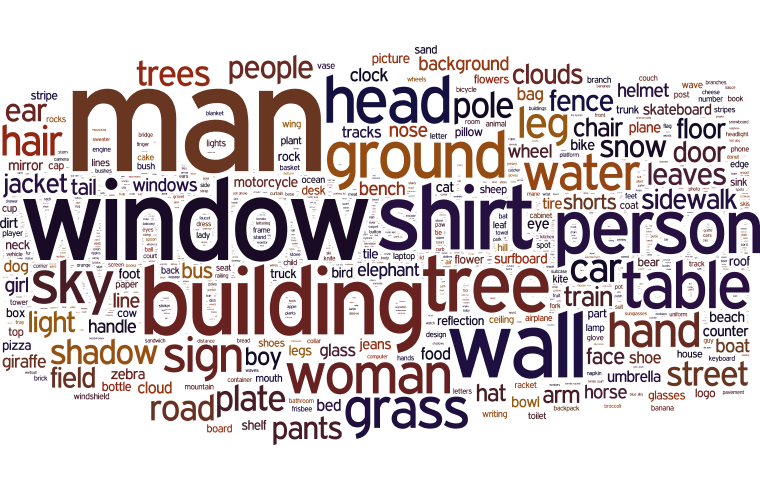}
\includegraphics[width=0.49\columnwidth]{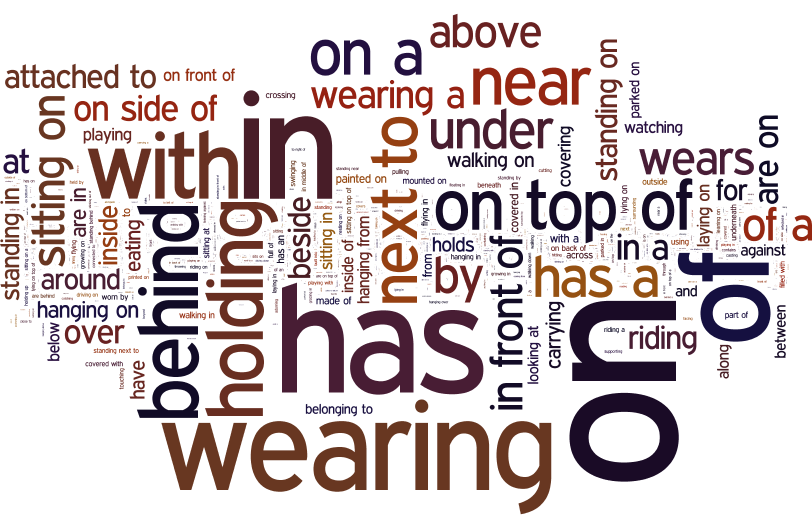}
\caption{Wordle images of object categories (left) and predicate categories (right) in Visual Genome (VG) dataset.}
\label{fig:vg_wordle}
\end{figure}

In our work, we use the proposed Intra-Hierarchical trees (IH-tree) to handle the ambiguous and noisy labels in Visual Genome (VG) dataset~\cite{krishna2017vg}.
Fig.~\ref{fig:vg_wordle} provides the wordle images\footnote{\url{http://www.wordle.net/}} to highlight the frequencies of object and predicate categories appeared in VG.
Bigger font sizes suggest higher frequencies.
The most frequent object is \emph{man} and the most common predicate is \emph{on}.

As shown in Fig.~\ref{fig:loss_hierarichy} in the main article, there are three levels in our Intra-Hierarchical trees, $\mathcal{H}_{o}$ and $\mathcal{H}_{p}$.
The class numbers in each level in our experiments are shown in Tab.~\ref{tb:ihtree_clsnum}.
The labels in $\mathcal{H}^{0}$ are the source labels in the datasets.
In VG, there are only $578$ classes of objects after clustering the labels in $\mathcal{H}^{2}$ by semantic similarity, which is much fewer than the original $5319$ classes.
Actually, there are nearly the same number of classes about verb-based and preposition-based predicates in $\mathcal{H}^{2}$ both in the VG and VRD datasets.
Since the annotation in VRD dataset is clean enough, we remove the intermediate layers $\mathcal{H}^{1}_{o}$ and $\mathcal{H}^{1}_{p}$ that aiming at reducing the label ambiguity and annotation noise.
%

\begin{table}
\caption{Number of classes in each layer in the Intra-Hierarchical Trees on VG and VRD datasets.}
\label{tb:ihtree_clsnum}
\centering
\begin{tabular}{M{2.0cm}|M{1.2cm} M{1.2cm} M{1.0cm}|M{1.2cm} M{1.0cm} M{1.0cm} M{1.0cm}} 
\hline
\multirow{2}{*}{Datasets} & \multicolumn{3}{c|}{\#Object} & \multicolumn{4}{c}{\#Predicate} \\\cline{2-8}
& $\mathcal{H}^{0}$ & $\mathcal{H}^{1}$& $\mathcal{H}^{2}$& $\mathcal{H}^{0}$ & $\mathcal{H}^{1}$& $\mathcal{H}^{2-1}$& $\mathcal{H}^{2-2}$\\
\hline
\hline
VG\cite{krishna2017vg}& 5,319 & 3,450 & 578 & 1,957 & 993 & 556 & 462\\
VRD\cite{lu2016lp}& 100 & - & 72 & 70 & - & 56 & 54\\
\hline
\end{tabular}
\vspace{0.1cm}
\end{table}

\begin{table}[t]
\centering
\caption{Performance drop on Visual Genome dataset by different unbalance loss weights compared to an equal weight applied to Zoom-Net.}
\label{tab:weights_result}
\footnotesize
\begin{tabular} {M{4.3cm}| M{1.5cm} M{1.5cm} M{1.5cm} M{1.5cm}}
\hline
 Performance drop & Sub. Acc@1 & Pred. Acc@1 & Obj. Acc@1 & Rel. Acc@1 \\
\hline
$\alpha=\gamma=1, \beta=0.5$ & 0.26 & 0.34 & 0.16 & 0.14\\
$\alpha=\gamma=1, \beta=2$ & 0.38 & 0.49 & 0.14 & 0.17\\
\hline
$\alpha=\gamma=1, \beta=0.1$ & 0.27 & 4.04 & 0.02 & 1.10\\
$\alpha=\gamma=1, \beta=10$ & 7.32 & 1.03 & 7.76 & 3.35\\
\hline
\end{tabular}
\vspace{0.15cm}
\end{table}

\subsection{Ablation Study}
\label{subsec:moreablation}

\subsubsection{Sensitivity to loss weights of multi-tasks.}
\label{subsubsec:loss_weight}
The parameters $\alpha,\beta,\gamma$ in Sec.~\ref{sec:ihtree} in the main body are to balance the scales of losses from three branches, so as to ensure balanced influences from \emph{subject, object} and \emph{relationship} during training.
Since these branches are evenly interacted with each other through the feed-forward pass, the back-propagated gradients from any loss can update the network parameters in other branches, thus a slight variance of weights for different losses will not have dominant effect on the training.
Tab.~\ref{tab:weights_result} shows the performance drop of using different scales of loss weights compared to equal weights. It is reasonable that the model may be sensitive to a large scale difference between \textit{predicate} ($\beta$) and \textit{subject/object} ($\alpha$/$\gamma$), while the small scale changes will not influence the results much.

\vspace{0.05cm}
\noindent\textbf{Computational time per image.} 
The computational cost of each component of Zoom-Net is listed in Tab.~\ref{tab:time}. The experiments are conducted on a single TITAN X GPU. A single SCA-M module (e.g.~after conv4\_3) only costs an additional 0.02s which make the whole framework efficient. If involving multiple SCA-M modules (e.g.~after conv3\_3), it may cause fewer shared layers and more time costs.

\begin{table}[t]
\centering
\vspace{0.25cm}
\caption{Computational time per image on Visual Genome dataset.}
\label{tab:time}
\vspace{-0.15cm}
\small
\begin{tabular} {M{3.5cm}| M{1.5cm} M{1.5cm} M{1.5cm} M{3.0cm}}
\hline
Net & A-M & CA-M & SCA-M & multi-SCA-Ms \\
\hline
Time (s/image)& 0.035 & 0.055 & 0.056 & 0.191 \\
\hline
\end{tabular}
\end{table}

\subsection{More Experiment Results of Zoom-Net}
\label{subsec:more_exp}

In this section, we show additional qualitative results on the VG dataset.
The experiment settings and details can be found in Sec.~\ref{sec:experiments_vg} in the main body.


\subsubsection{Scene Graph Generation}
\label{subsubsec:scenegraph}

\begin{figure}[t]
\centering
\includegraphics[width=1\columnwidth]{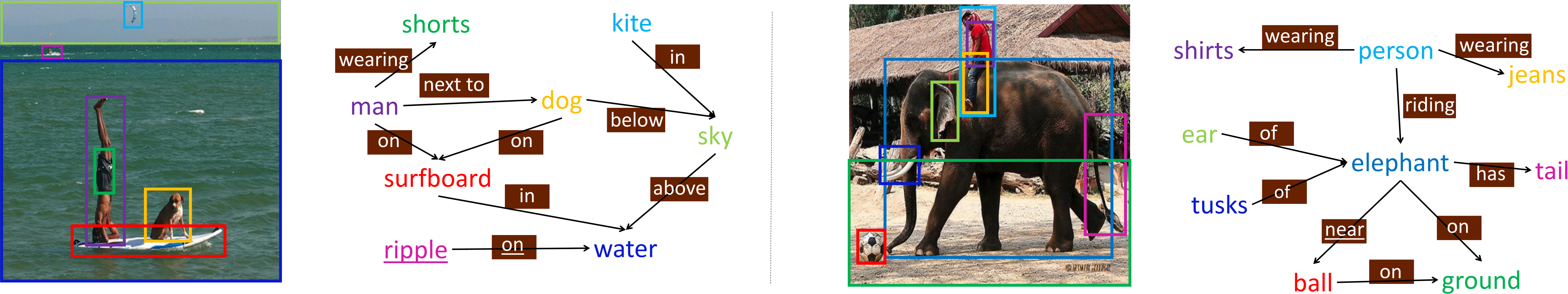}
\caption{Two exemplar scene graphs generated by the proposed Zoom-Net. The underlined words are wrong predictions of the subject, predicate or object. The font color of the object category in the scene graph is related to the rectangle color in the corresponding image left.
}
\label{fig:exp_graph}
\end{figure}

Scene graph generation can serve as the basis for a number of tasks, e.g. visual question answering~\cite{wu2017visual} and image retrieval~\cite{johnson2015image}
The proposed Zoom-Net can also perform well on scene graph generation.
The task here is to generate a directed graph for an image that captures objects and their relationships.
Fig.~\ref{fig:exp_graph} illustrates two scene graphs generated by the proposed Zoom-Net. The reported excellent performances come from the proposed effective and efficient visual relationship recognition.

\subsubsection{Zero-shot Relationship Recognition.}
\label{subsubsec:zeroshot}

Owing to the long tail distribution of relationship labels in VG dataset, even though each single object or predicate category can be guaranteed to appear both in the training and testing sets, it is hard to assure the distribution of their combination~(\textit{i.e.},~tuple relationship). This results in a zero-shot relationship recognition problem. 
A couple of examples are shown in the first row of Fig.~\ref{fig:exp_qualitative_result_app}(c). They~(\textit{e.g.},~\tuple{water}{in}{window} and \tuple{vase}{on}{head}) are not in the training set. Compared to the reference methods, these unseen relationships can be well inferred by our model using similar relationships~(\textit{e.g.},~\tuple{person}{in}{window} and \tuple{hat}{on}{head}) learned from the training set.

\subsubsection{Additional Results on Visual Genome}
\label{subsubsec:exp_result}

\begin{figure*}[t]
\centering
\includegraphics[width=1.0\linewidth]{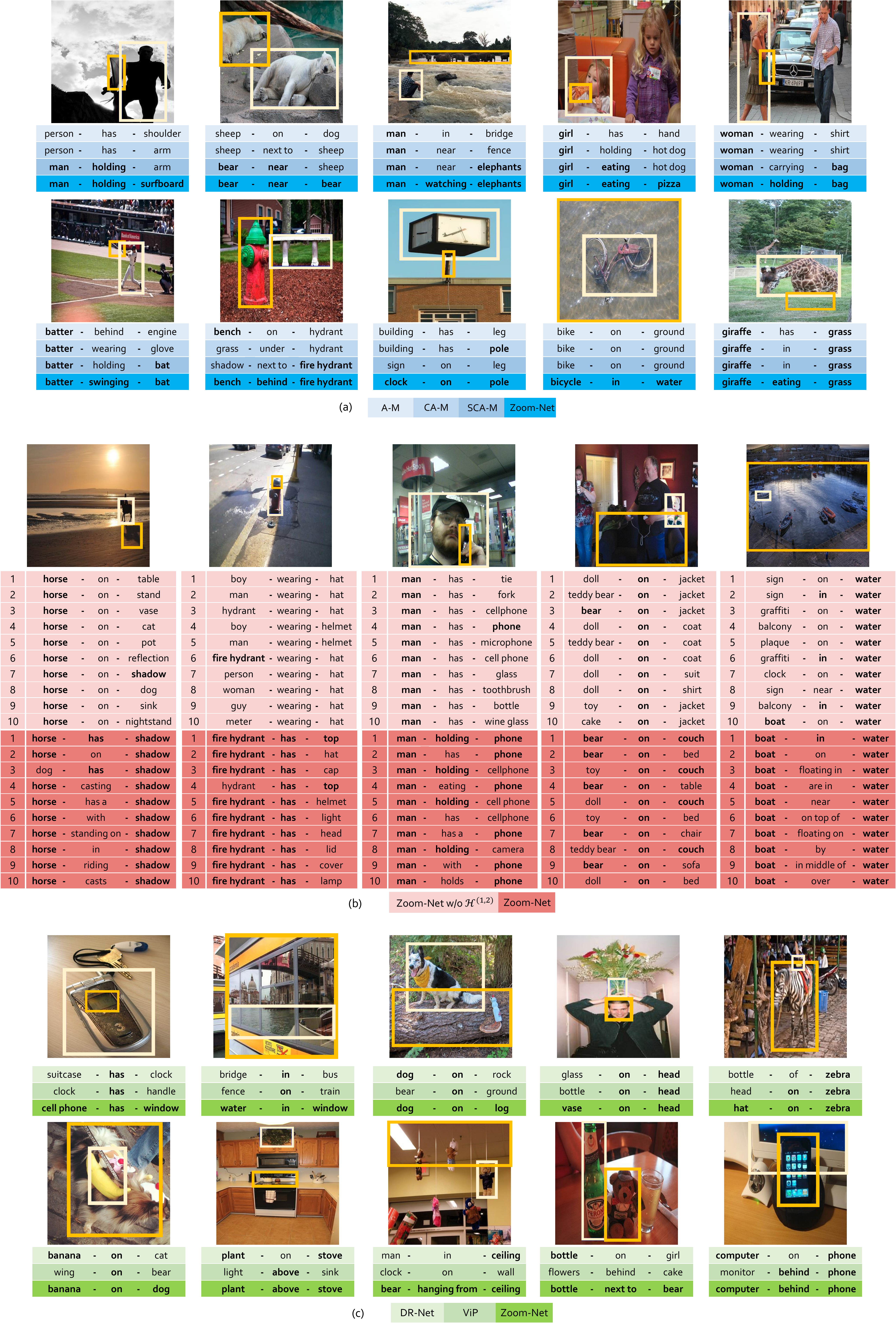}
\caption{More qualitative results on VG dataset. (a) Comparison among different module configurations. (b) Results by discarding IH-trees. (c) Comparison between Zoom-Net and state-of-the-art methods. (a) and (c) show Top-$1$ prediction results while (b) provides Top-10 results for each method. The ground truth are in bold font.
}
\label{fig:exp_qualitative_result_app}
\vspace{-0.3cm}
\end{figure*}

The additional qualitative comparisons are in visualized Fig.~\ref{fig:exp_qualitative_result_app}.
Firstly, we compare the results among the different module configurations of the proposed Zoom-Net.
Then we show the Top-10 triple relationship prediction results of Zoom-Net with and without IH-tree in Fig.\ref{fig:exp_qualitative_result_app}(b).
Finally, Fig.\ref{fig:exp_qualitative_result_app}(c) shows the excellent performance of Zoom-Net, compared with the state-of-the-art methods, DR-Net~\cite{dai2017drnet} and ViP~\cite{li2017vip}. The details are depicted in Sec.~5.1 and Sec.~5.2 in the main body.

\end{document}